\documentclass[3p]{elsarticle}
\usepackage{lipsum}
\makeatletter
\def\ps@pprintTitle{%
 \let\@oddhead\@empty
 \let\@evenhead\@empty
 \def\@oddfoot{}%
 \let\@evenfoot\@oddfoot}
\makeatother

\usepackage{bm}
\usepackage{mathtools}
\usepackage{multirow}
\usepackage{amsfonts,amssymb}
\usepackage{subcaption}
\usepackage{lineno,hyperref}
\usepackage{pifont}
\usepackage{amssymb}
\usepackage{graphicx}
\usepackage{subcaption}
\usepackage{float}
\usepackage{color,soul}
\newcommand{\cmark}{\ding{51}}%
\newcommand{\xmark}{\ding{55}}%
\usepackage[table]{xcolor}

\begin{document}

\begin{frontmatter}

\title{Training and Inference for Integer-Based \\ Semantic Segmentation Network}


\author[label1]{Jiayi Yang}
\author[label2]{Lei Deng}
\author[label3]{Yukuan Yang}
\author[label2]{Yuan Xie}

\author[label3]{Guoqi Li\corref{mycorrespondingauthor}}
\cortext[mycorrespondingauthor]{Corresponding author}
\ead{liguoqi@mail.tsinghua.edu.cn}

\address[label1]{International School, Beijing University of Posts and Telecommunications, Beijing 100876, China.}
\address[label2]{Department of Electrical and Computer Engineering, University of California, Santa Barbara, CA 93106, USA.}
\address[label3]{Department of Precision Instrument, Center for Brain Inspired Computing Research, Tsinghua University, Beijing 100084, China.}

\begin{abstract}
Semantic segmentation has been a major topic in research and industry in recent years. However, due to the computation complexity of pixel-wise prediction and backpropagation algorithm, semantic segmentation has been demanding in computation resources, resulting in slow training and inference speed and large storage space to store models. Existing schemes that speed up segmentation network change the network structure and come with noticeable accuracy degradation. However, neural network quantization can be used to reduce computation load while maintaining comparable accuracy and original network structure. Semantic segmentation networks are different from traditional deep convolutional neural networks (DCNNs) in many ways, and this topic has not been thoroughly explored in existing works. In this paper, we propose a new quantization framework for training and inference of segmentation networks, where parameters and operations are constrained to 8-bit integer-based values for the first time. Full quantization of the data flow and the removal of square and root operations in batch normalization give our framework the ability to perform inference on fixed-point devices. Our proposed framework is evaluated on mainstream semantic segmentation networks like FCN-VGG16 and DeepLabv3-ResNet50, achieving comparable accuracy against floating-point framework on ADE20K dataset and PASCAL VOC 2012 dataset.
\end{abstract}


\end{frontmatter}

\section{Introduction}
Semantic Segmentation has been a major research focus since the beginning of this field. Recently, the thriving of deep learning inspired researchers to handle this task with deep neural networks\cite{wang2017gated}\cite{long2015fully}, and many of them has outperformed traditional algorithms\cite{lateef2019survey}. Recent models such as PSPNet\cite{zhao2017pyramid}, DeepLab series\cite{chen2014semantic}\cite{chen2017deeplab}\cite{DBLP:journals/corr/ChenPSA17}\cite{chen2018encoder} have achieved impressive results on public datasets. However, compared with object classification task, semantic segmentation in deep learning suffers from huge computation cost and storage space because of its pixel-wise prediction. For instance, consider a ResNet50 DCNN and a DeepLab-ResNet50 segmentation network: it takes around 2 ms to train one image in ResNet50 and 25 ms for DeepLab with the same input size $224 \times 224$ on an Nvidia Titan V GPU. To generate a full-size semantic segmentation prediction, it takes much longer than to expect a single classification result in DCNN. This property makes shifting from traditional segmentation to real-time segmentation suffering.

Regarding this problem, recent work on real-time semantic segmentation networks \cite{paszke2016enet} \cite{zhao2018icnet} \cite{yu2018bisenet} often design new network structure to trade off accuracy for inference speed. To achieve higher efficiency while maintaining similar accuracy calls for other methods than to design new network structures. Recent works on neural network quantization managed to lower the bit-width of dataflow while maintaining accuracy. The quantized networks restrict parameters and computation to lower bits, and simulate full precision training and inference with discrete dataflow.

\begin{figure}
\setcounter{figure}{0} 
 \centering
 \includegraphics[width=0.6\linewidth]{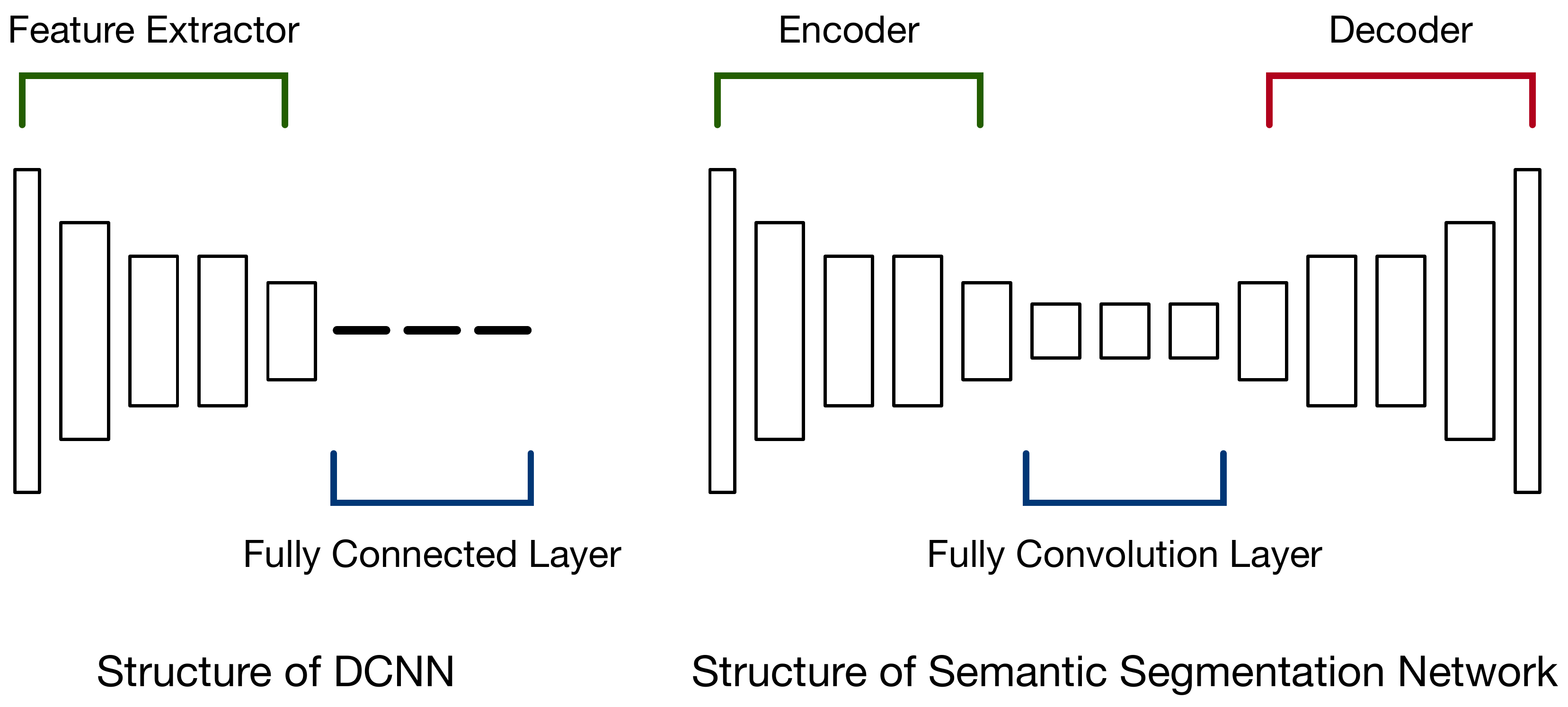}
 \caption{Difference of general structure between DCNN and semantic segmentation networks.}
 \label{struc_comp}
\end{figure}

However, major works on network quantization like BNN \cite{DBLP:journals/corr/CourbariauxB16}, WAGE \cite{wu2018training} mainly explore quantization on DCNN designed for object classification. Due to the complexity of pixel-wise prediction, semantic segmentation tasks have to use deeper networks. This causes some quantization methods in DCNN with fewer layers not suitable for semantic segmentation. Furthermore, existing works of quantization on semantic segmentation task \cite{tang2018quantized} \cite{tang2019towards} do not fully quantize all the parameters in the network, making it hard to be implemented on integer-based deep learning chip, or to generalize to other existing models that are not dedicatedly quantized.

The general structure comparison between DCNN and semantic segmentation network is shown in Figure \ref{struc_comp}. First, it removes the fully connected layers from the picture and replaces it with convolution layers. Second, as DCNN is the encoder of the semantic segmentation network, there also exists a decoder structure that recovers the feature map dimension. Third, segmentation network weights are initialized differently. The encoder is usually initialized with pre-trained weight from DCNN, and the decoder part is initialized differently in various networks. Therefore, to quantize semantic segmentation network is not so intuitive as implementing quantization frameworks on DCNN.

Hence, we propose a new framework that quantizes deep semantic segmentation network into integer-based dataflow, constraining nodes into low bit discrete space. In addition to weight ($W$), activation ($A$), error ($E$), gradient ($G$), and update ($U$), we address batch normalization \cite{DBLP:journals/corr/IoffeS15} to train deeper models, which is often neglected or replaced in previous works on quantization. While batch normalization helps the training of deeper network and speeds up convergence, it also contributes to a large portion of computation on run time, and the nonlinearity introduced by square and root operations makes it difficult to be quantized. Fortunately, L1-norm batch normalization (L1BN) \cite{8528524} is proven to be mathematically equivalent to L2-norm batch normalization (L2BN) but demonstrates stronger linearity. Consequently, we quantized L1BN in our network instead of L2BN, therefore endowing it the ability to run on integer-based hardware, which falls short on computing square root operations.

Our proposed framework is evaluated on ADE20K dataset\cite{zhou2016semantic} and PASCAL VOC 2012 datasets\cite{pascal-voc-2012} on mainstream segmentation network FCN and DeepLabv3. We achieve comparable accuracy compared with full precision networks when constraining major dataflow into 8 bits integers, striking a balance between bit-width and performance. We conduct experiments and analyses on quantized semantic segmentation networks from different perspectives (e.g. bit-width, quantization details), providing some insights for further adaptation on other segmentation networks.

In summary, our contributions are: 
\begin{itemize}
\item We propose a framework for semantic segmentation network that constrains the major dataflow of training and inference to 8-bit integers.
\item Our framework achieves comparable performance as the full-precision network. Designed for general segmentation network, our framework leaves room for application on other models for further work.
\item We perform a thorough analysis and experiment of different factors impacting of performance of semantic segmentation network.
\end{itemize}

\section{Related Work}
Many models have been developed since the development of Fully Convolutional Network. In this work, we experiment on two classic networks that can represent most of the mainstream networks used today.

\subsection{Semantic Segmentation}
Many models have been developed since the development of Fully Convolutional Network. In this work, we experiment on two classic networks that can represent most of the mainstream networks used today.

\textbf{Fully Convolutional Network:} Fully convolutional network represents the broad class of networks for semantic segmentation. It replaces fully connected layers in traditional CNN with convolution layers and appends a decoder that recovers feature map resolution to produce pixel-wise semantic segmentation prediction. Although it is more efficient to predict one label, as its output is a pixel-wise label, the computation overhead is huge. In this paper, we use the classic FCN8s with VGG16 \cite{simonyan2014very} as it's encoder in the experiments.

\textbf{DeepLab:} DeepLab architecture has been one of the most accurate models for the task of semantic segmentation since the work of FCN. It mainly contains a DCNN backbone appended with an atrous spatial pyramid pooling(ASPP) module. DenseCRF was removed from the framework to maintain simplicity. Most recent DeepLabv3\cite{DBLP:journals/corr/ChenPSA17} and DeepLabv3+ have achieved state-of-the-art accuracy on public datasets.

\subsection{Network Quantization}
Neural network quantization \cite{courbariaux2015binaryconnect} \cite{rastegari2016xnor} \cite{deng2018gxnor} \cite{yang2020training} is the approach of trying to reduce the model size and accelerate computation by reducing the bit-width of the operands in the network. In recent works, \emph{Wu et al.}\cite{wu2018training} proposed WAGE to discretize parameters in both training and inference. They identify weight ($W$), activation ($A$), error ($E$), and gradient ($G$) in forward and backward propagation, and constrain them to low bit-width integers. To simulate BN, they replaced it with a layer-wise constant scaling factor. However, WAGE's experiments were carried on shallow CNNs, and the performance on deeper networks are not satisfying. Other works like 8b Training\cite{wang2018training} and FX Training\cite{sakr2018per} push quantization frameworks to deeper CNNs, though BN is rarely mentioned in these frameworks. To sum up, despite the thriving research of quantization in traditional CNN, there is still room to quantize in the dataflow of neural networks. Moreover, quantizing semantic segmentation networks is not as intuitive due to the structural and detailed differences with DCNN.

\section{Network Quantization}
Quantization method and quantization object are two important factors when considering network quantization. Different objects may use different methods depending on their data distribution and property. In this section, we first identify the quantization methods. Then, we introduce the entire framework in the order of forward pass and then back pass. 

\subsection{Quantization Methods and Distribution Analysis}
\subsubsection{Uniform Quantization}
Uniform quantization is the basis of other quantization methods. The minimum quantization distance is governed by
\begin{equation}\label{del}
\Delta(k) = 2^{1-k}
\end{equation}
where $k$ is the quantization bit-width.
Uniform quantization is similar to ADCs used in signal processing, which deterministically maps the floating-point values into nearest discrete state. Uniform quantization is defined as
\begin{equation}\label{UQ}
UQ(x,k)= Clip(\Delta*\lfloor\frac{x}{\Delta} + \frac{1}{2}\rfloor,-1+\Delta,1-\Delta)
\end{equation}
where $x$ is the quantization target and $\Delta$ is the quantization distance with respect to parameter $k$. $Clip(\cdot)$ function clips the quantized value inside the quantization range between $[-1+\Delta,1-\Delta]$.

\begin{figure}
\setcounter{figure}{1} 
 \centering
 \includegraphics[width=0.5\linewidth]{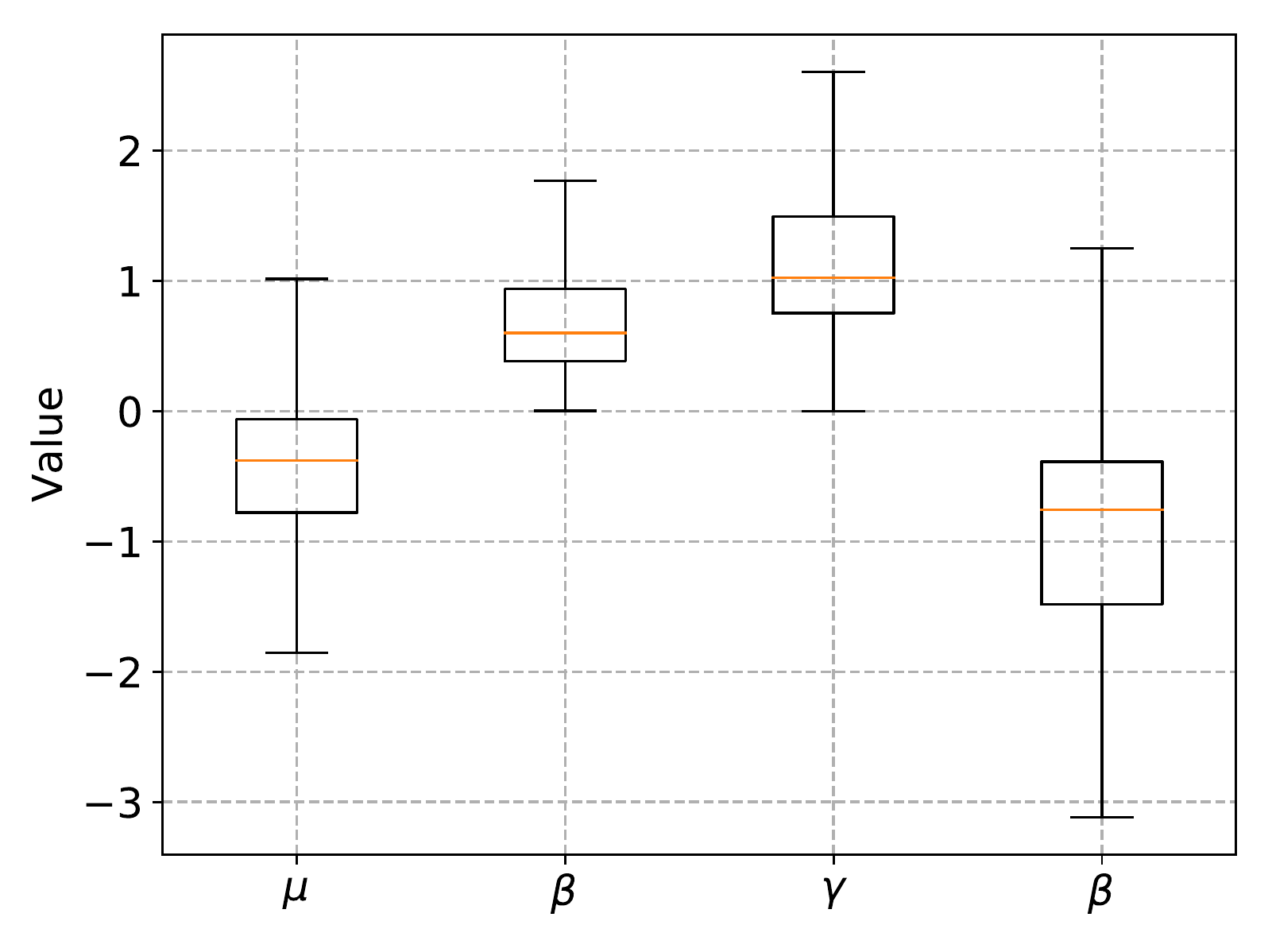}
 \caption{Quartile box graph of $\sigma$, $\mu$, $\gamma$ and $\beta$. The majority of value lies between $[-2, 2]$.}
 \label{mvgb}
\end{figure}

Quantization objects for semantic segmentation networks without BN include weight ($W$), activation ($A$), error ($E_1$), gradient ($G$), and update ($U$). $E_1$ is defined as the gradient of activation, and $G$ is the gradient of weight. For networks with BN, variance ($\sigma$), mean ($\mu$), normalized output ($\hat{x}$), scale ($\gamma$), shift ($\beta$), and error ($E_2$) is included. The additional $E_2$ is defined as the gradient of normalized output in BN.

Figure \ref{mvgb} shows the distribution of $\sigma$, $\mu$, $\gamma$ and $\beta$. Experiments reveal that these values lie stably inside the range of$[-2,2]$ throughout the training process, and the therefore we perform constant scaling of $2$ and $2^{-1}$ before and after uniform quantization to fit the quantization range of $[-1+\Delta,1-\Delta]$.

\subsubsection{Scale Quantization}\label{SQsection}
However, quantization for $W$, $A$, $E_1$, $E_2$, $G$ and normalized output $X$ is not as straightforward. Figure \ref{object_distri} illustrates the distribution of these objects in full-precision DeepLabv3-ResNet50 network during training. Here we identify two failure modes considering uniform quantization $UQ(\cdot)$: 
\begin{itemize}
\item \textbf{First} failure mode is when the distribution is concentrated between the smallest quantization step $\Delta(k)$.
\item \textbf{Second} failure mode is when considerable portion of weights are clipped by quantization boundary. 
\end{itemize}
As all the quantization object distribution mentioned in Figure \ref{object_distri} except $W$ belongs to one of these two modes, the framework suffers performance degradation when using uniform quantization.

\begin{figure}
\setcounter{figure}{2}
 \centering
 \includegraphics[width=\linewidth]{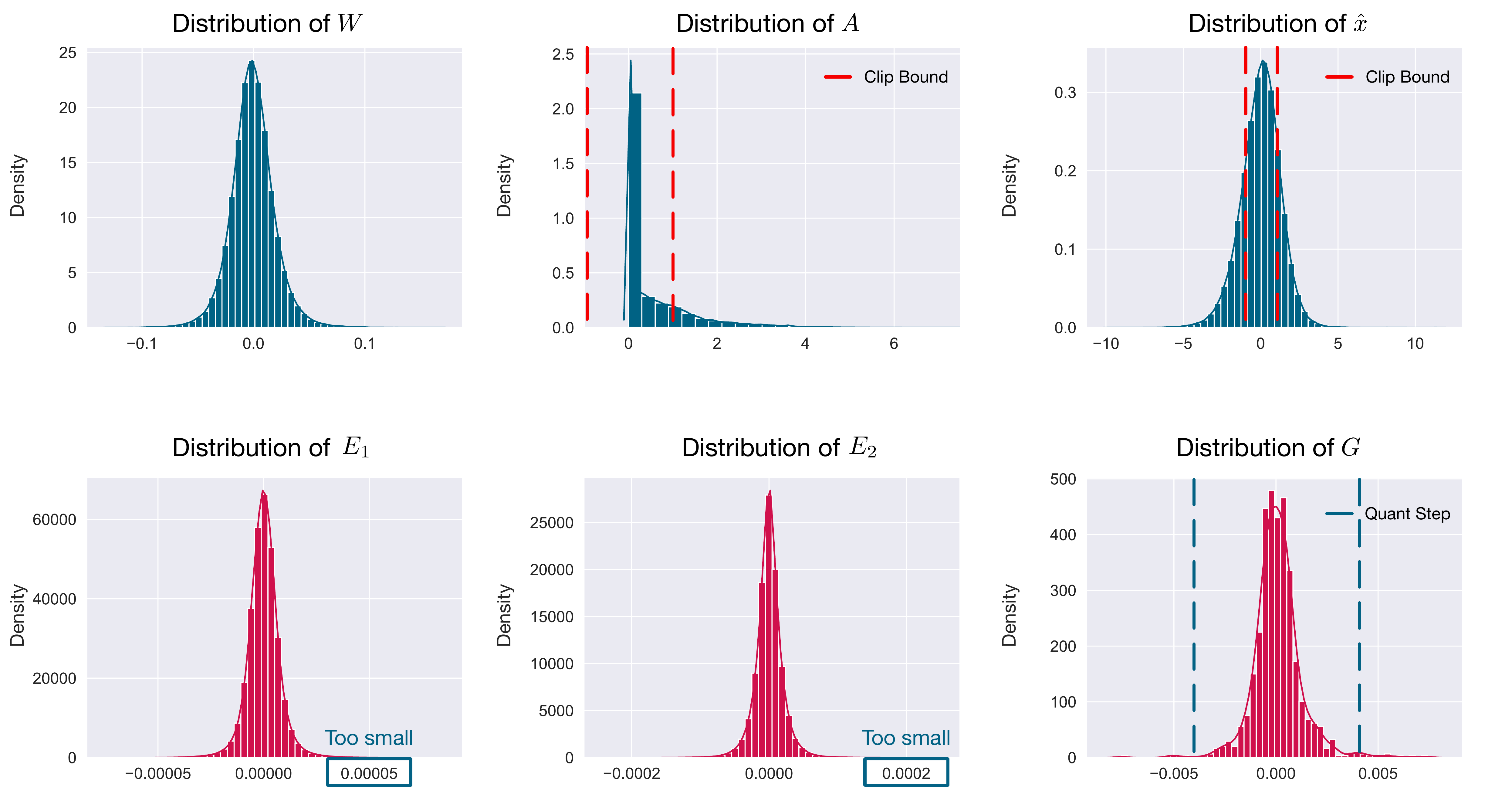}
 \caption{Distribution comparison of various quantization objects. The distribution in blue represents forward propagation dataflow, and the distribution in red represents backward propagation.}
 \label{object_distri}
\end{figure}

Scale quantization can be applied to escape from these two failure modes. An important component in scale quantization is the dynamic sacaling factor. Various scaling factors have been used in previous works, but our experiments show that the naive dynamic scale factor
\begin{equation}\label{ScaleFactor}
Scale(x) = max(\left|x\right|)
\end{equation}
does not differ with in terms of semantic segmentation network performance. For any object $x$, most of the normalized value in $\frac{x}{Scale(x)}$ will lie in the quantization range $[-1+\Delta,1-\Delta]$. The scale quantization equation is given as
\begin{equation}\label{ScaleQ}
SQ(x,k) = Scale(x) * UQ(\frac{x}{Scale(x)},k)
\end{equation}
where $x$ will be scaled by $Scale(x)$ before quantization to fit data range and scaled back after quantization to preserve its original order of magnitude.

\textbf{$\bm{A}$ and $\bm{\hat{x}}$ distribution}. The distribution of activation and normalized output is far different from weight. They both fall into the \textbf{second} mode by exceeding the quantization range. The values clipped by function $Clip(\cdot)$ will result in information loss that affects training during loss calculation. Hence, we use scale quantization for activation and normalized output in BN to ensure network convergence.

\textbf{$\bm{E_1}$ and $\bm{E_2}$ distribution}. Both error objects fall into the \textbf{first} failure mode where the order of magnitude is smaller than the quantization step $\Delta$. Suppose that bit-width $k_E=8$, the quantization step $\Delta(k_E)\approx0.008$ far exceeds the order of magnitude of error in $[10^{-5},10^{-4}]$ range. This will result in error propagation with all the value rounded to zero. Therefore we also apply scale quantization given in Equation (\ref{ScaleQ}) for error in the dataflow.

\textbf{$\bm{G}$ distribution}. Gradient have similar distribution when compared to error, but through experiments, we found that network performance drops noticeably when rounding to around 8 bit. This suggests gradient is more sensitive to bit-width and preservation of small values. To solve this problem, we round gradients stochastically in place of $UQ(\cdot)$.

\subsubsection{Stochastic Quantization}
Suppose we have a properly scaled quantization object $x$, the definition of stochastic quantization is governed by
\begin{equation}\label{SQ}
RQ(x,k)= Clip(\Delta*Round(\frac{x}{\Delta}), -1+\Delta,1-\Delta)
\end{equation}
where $\Delta$ denotes the quantization step according to $k$ defined in Equation (\ref{del}). $RQ(\cdot)$ largely resembles uniform quantization, except that we round the values stochastically using
\begin{equation}\label{SR}
Round(x) =
\left\{  
 \begin{array}{c}
 \lfloor x\rfloor \ \ \ \ \ \ \ \ P =\lceil x\rceil - x\\
 \\
 \lfloor x\rfloor + 1 \ \ \ P =x - \lfloor x\rfloor\\
 \end{array}
\right
..
\end{equation}

The floating-point number $x$ will be rounded to its two neighbor integers stochastically, and the probability $P$ correlates to the distance to that integer linearly. The expectation of stochastic rounding $\mathbb{E}(Round(x))=x$, resulting in no expected error compared to deterministic rounding. Further, stochastic rounding feeds additional probability information into the dataflow, and gradients inside the smallest quantization step can also be rounded to non-zero values. Details of gradient quantization will be covered in Section \ref{Gradient Section}.

\begin{figure}
\setcounter{figure}{3}
 \centering
 \includegraphics[width=\linewidth]{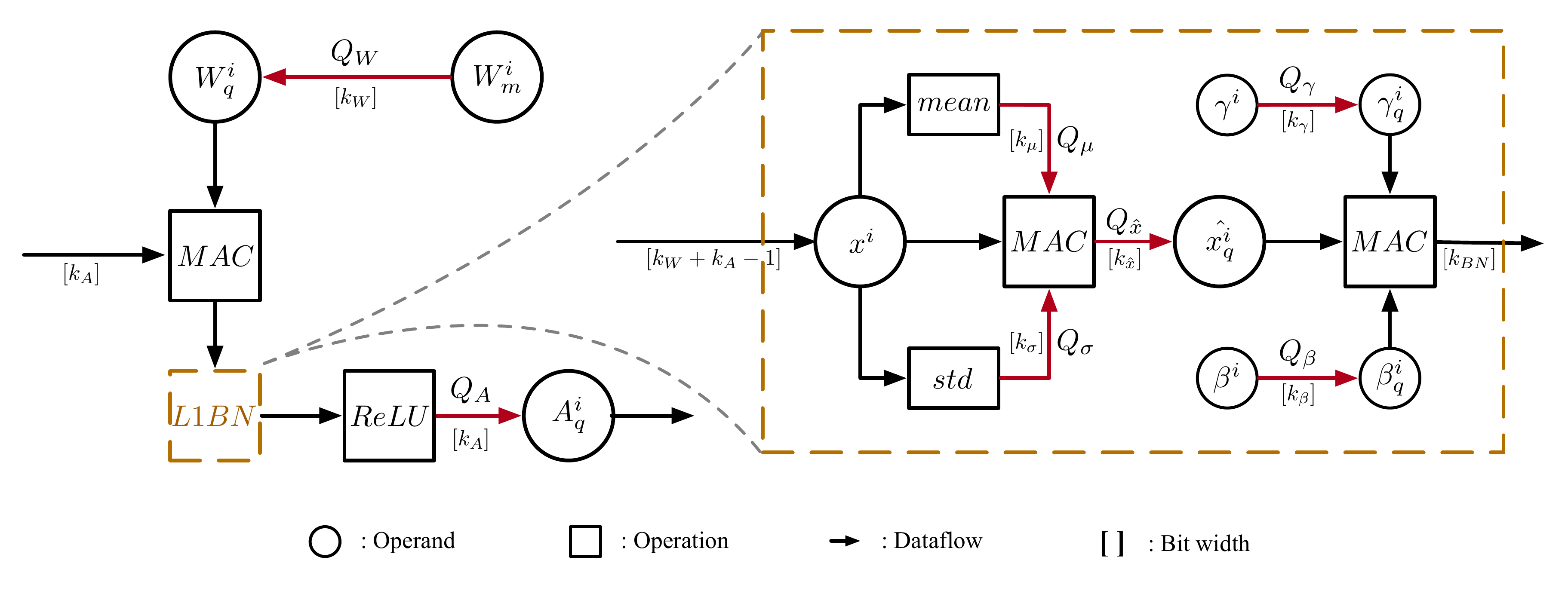}
 \caption{Dataflow and quantization framework in forward propagation. The red arrows indicate the quantization and dataflow of discrete parameters. L1BN module is removable from the quantization framework.}
 \label{forward}
\end{figure}

\subsection{Forward Propagation}
\subsubsection{Weight Quantization}
Initialization of weight in a semantic segmentation network is different from DCNN. The encoder structure is initialized with full-precision weight trained from object classification tasks like ImageNet. Fully convolution layers are initialized with MSRA initialization\cite{he2015delving}, and the decoder is initialized as bilinear upsampling filter\cite{long2015fully}. Since we train the network with discrete dataflow, initialized weights must also be in a discrete state by 
\begin{equation}\label{WQ0}
W_m^i = UQ(W^i,k_U)
\end{equation}
with weight update bit-width $k_U$. We also discover that the standard initialization approach converges well for fully convolution layers and decoder structure, and other initialization methods do not contribute to noticeable speed up in convergence nor higher accuracy. \\
\indent During forward propagation, weights are directly quantized with uniform quantization
\begin{equation}\label{WQ1}
W_q^i = Q_W(W_m^i)= UQ(W_m^i,k_W)
\end{equation}
where $k_W$ denotes the quantization bit-width of weights, and $W_m^i$ denotes the master weights stored in the network. Note that master weights are initialized and stored with update bit-width $k_U$, covered with details in Section \ref{Gradient Section}. As shown in Figure \ref{forward}, the master weights stored with $k_U$ bit is first quantized to $k_W$ bit by $Q_W$ before feeding into the convolution layer. \\

\subsubsection{L1BN Quantization}\label{L1BNQ}
Batch normalization (BN), serving as an important normalizing method, is not often addressed in existing quantization frameworks. However, we cannot ignore BN if we want to quantize the entire dataflow. Traditional BN with L2-norm can be summarized by
\begin{equation}\label{BN1}
\hat{x^i} = \frac{x^i - \mu_\mathcal{B}}{\sqrt{\sigma_\mathcal{B} + \epsilon}}, \ y^i = \gamma \hat{x^i} + \beta
\end{equation}
where the former normalizes the distribution, and the later recovers some representation ability lost from the normalization operation; $\epsilon$ demotes a small value added for numerical stability; $\mu_\mathcal{B}$ and $\sigma_\mathcal{B}$ represents the mean and variance calculated from the mini-batch $\mathcal{B}$, where
\begin{equation}\label{BN2}
\mu_\mathcal{B} = \frac{1}{m}\sum_{i=1}^m x^i
\end{equation}
and
\begin{equation}\label{BN3}
\sigma_\mathcal{B} = \sqrt{\frac{1}{m}\sum_{i=1}^m(x^i - \mu_\mathcal{B})^2}
\end{equation}
respectively. However, the root and square operations in Equation (\ref{BN1}) and (\ref{BN3}) brings strong nonlinearity, making it hard to quantize the BN dataflow and to implement on low bit-width hardware. Fortunately, while original BN introduces root and square from the L2-norm, it's low order counterpart L1-norm variance demonstrates better linearity \cite{8528524}. Thus, $\hat{x^i}$ in Equation (\ref{BN1}) is replaced with
\begin{equation}\label{BN4}
\hat{x^i} = \frac{x^i - \mu_\mathcal{B}}{\sigma_\mathcal{B} + \epsilon}
\end{equation}
where $\mu_\mathcal{B}$ remains the same, and $\sigma_\mathcal{B}$ replaced with
\begin{equation}\label{BN5}
\sigma_\mathcal{B} = \frac{1}{m}\sum_{i=1}^m\left|x^i - \mu_\mathcal{B}\right|
\end{equation}
representing L1-norm variance instead of L2-norm variance.\\
\indent In L1BN quantization, we identify five quantization objects: $\mu_\mathcal{B}, \sigma_\mathcal{B}, \gamma, \beta$ and $\hat{x^i}$. The additional $\hat{x^i}$ has to be quantized due to the increase of bit-width in the normalization operation in Equation (\ref{BN1}). The quantization of BN can be described as
\begin{equation}
 \begin{gathered}
\mu_q = Q_\mu (\mu_\mathcal{B}) = 2*UQ(\frac{\mu_\mathcal{B}}{2},k_\mu)\\
\sigma_q = Q_\sigma(\sigma_\mathcal{B}) = 2*UQ(\frac{\sigma_\mathcal{B}}{2},k_\sigma)\\
\hat{x_q^i} = Q_{\hat{x}}(\hat{x^i}) = SQ(\hat{x^i},k_{\hat{x}})\\
\gamma_q = Q_\gamma(\gamma) = 2*UQ(\frac{\gamma}{2},k_\gamma)\\
\beta_q = Q_\beta(\beta) = 2*UQ(\frac{\beta}{2},k_\beta)
 \end{gathered}
\end{equation}
where $k_\mu, k_\sigma, k_\gamma, k_\beta$ and $k_{\hat{x}}$ denotes the bit-width of the quantization objects, respectively.
The dataflow of quantized BN is shown in the right part of Figure \ref{forward}, the bit-width of output is $k_{BN}$. Depending on the network structure, the L1BN quantization block can be removed without changing other components in the framework.

\subsubsection{Activation Quantization}
After the MAC operations of convolution or scaling in BN (depending on if the network implements BN or not), the precision increases. As a result, the bit-width of activation has to be limited before the input of the next convolution layer. Activation is governed by scale equation where
\begin{equation}\label{AQ1}
A_q^i = Q_A(A^i)= SQ(A^i,k_A)
\end{equation}
with $k_A$ representing the bit-width of activations.

\begin{figure}
 \centering
 \includegraphics[width=\linewidth]{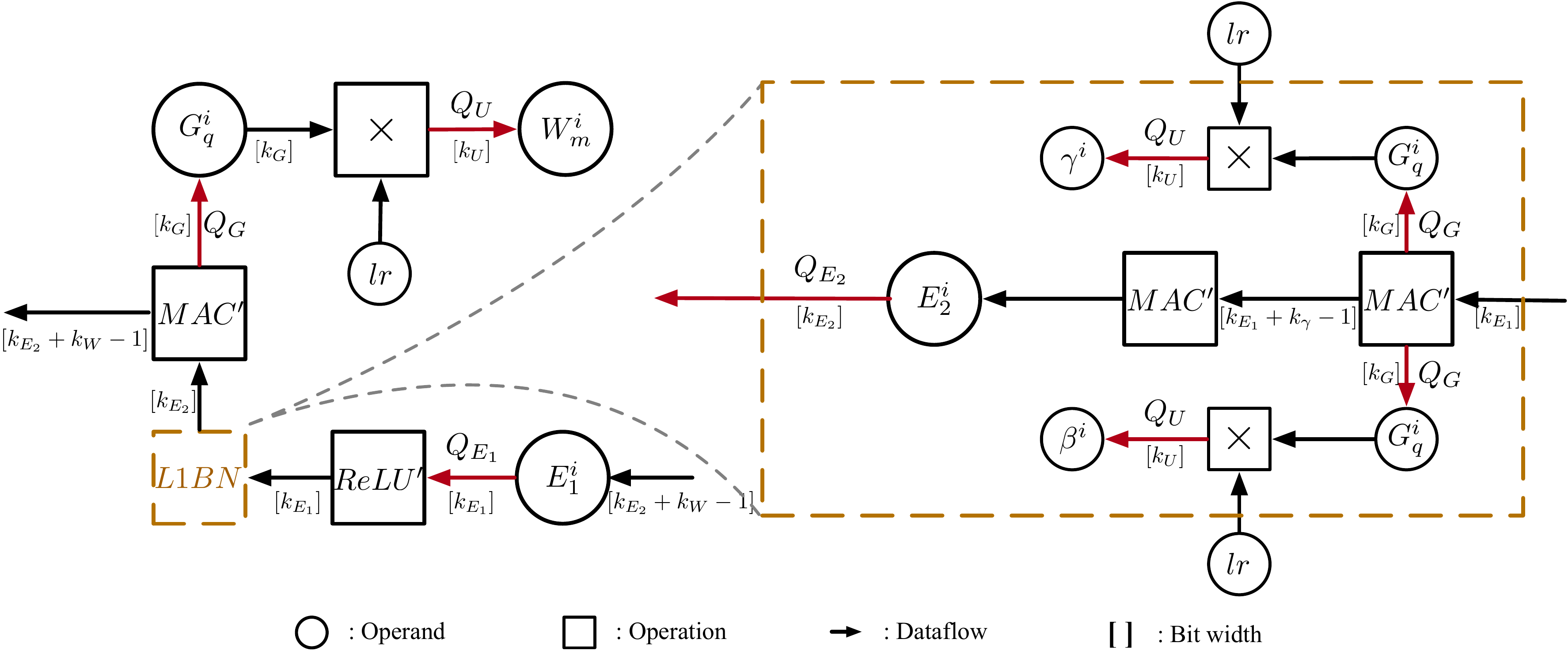}
 \caption{Dataflow and quantization framework in back propagation. The red arrows indicates the quantization and dataflow of discrete parameters. L1BN module is removable from the quantization framework.}
 \label{backward}
\end{figure}

\subsection{Backward Propagation}
\subsubsection{Error Quantization}
As mentioned earlier, scale quantization is used for objects like error to prevent the values from being zeroed out, then we have
\begin{equation}\label{EQ}
E_q^i = Q_E(E^i) = SQ(E^i,k_E)
\end{equation}
where $E^i$ is defined as the gradient of activation in each convolution and deconvolution layer $i$, denoted as $E_1^i$ in Figure \ref{backward}. It will be quantized by $Q_E$ with bit-width $k_{E_1}$ and then used for further calculations in the chain rule. For networks with BN, an additional $E_2^i$ should be quantized. The bit-width of error will increase according to the chain rule during backpropagation, so $E_2^i$ is further restricted by $Q_E$ to $k_{E_2}$, which is related to weight update.\\

\subsubsection{Gradient Quantization}\label{Gradient Section}
Gradients are quantized using the stochastic quantization method. Similar to uniform quantization, dynamic scaling has to be performed to avoid the \textbf{first} failure mode. However, while scale quantization scales back to the original order of magnitude after quantization, we identify that this does not necessarily apply to gradients. As the last step before weight update, gradients can be more sensitive to the order of magnitude. Hence, we give two versions of the gradient quantization method depending on the network structure. For a network that implements BN, gradients are quantized by
\begin{equation}\label{GQ1}
G_q^i=Q_G(G^i)= Scale(G^i)*RQ(\frac{G^i}{Scale(G^i)},k_G)
\end{equation}
where the original order of magnitude is preserved by $Scale(G^i)$. For a network that does not implement BN, gradients are quantized by
\begin{equation}\label{GQ2}
G_q^i=Q_G(G^i)= RQ(\frac{G^i}{Scale(G^i)},k_G)
\end{equation}
The difference between Equation ({\ref{GQ1}) and (\ref{GQ2}) is that the latter abandons the original magnitude. We expand the detailed analysis in Section \ref{backpropA}.

\subsubsection{Update Quantization}
Weight update directly reflects on master weights saved in the network, quantized by
\begin{equation}\label{UQ2}
\Delta W^i = Q_U(G_q^i,k)= UQ(G_q^i*lr,k_U)
\end{equation}
where gradient $G_q^i$ is multiplied by quantized learning rate $lr$, and $lr$ gradually decrease discretely during the entire training process.\\

The update takes the gradient and scales it by the learning rate. The product is quantized by $Q_U(\cdot)$ with bit-width $k_U$, returning the quantized weight update. It is responsible for the update of weights, and $\gamma$, $\beta$ from the BN layer. Experiments show that the update quantization bit-width $k_U$ should always be higher than $k_W$ for the network to converge.

\begin{figure}
 \centering
 \begin{subfigure}{0.5\linewidth}
 \includegraphics[width=\linewidth]{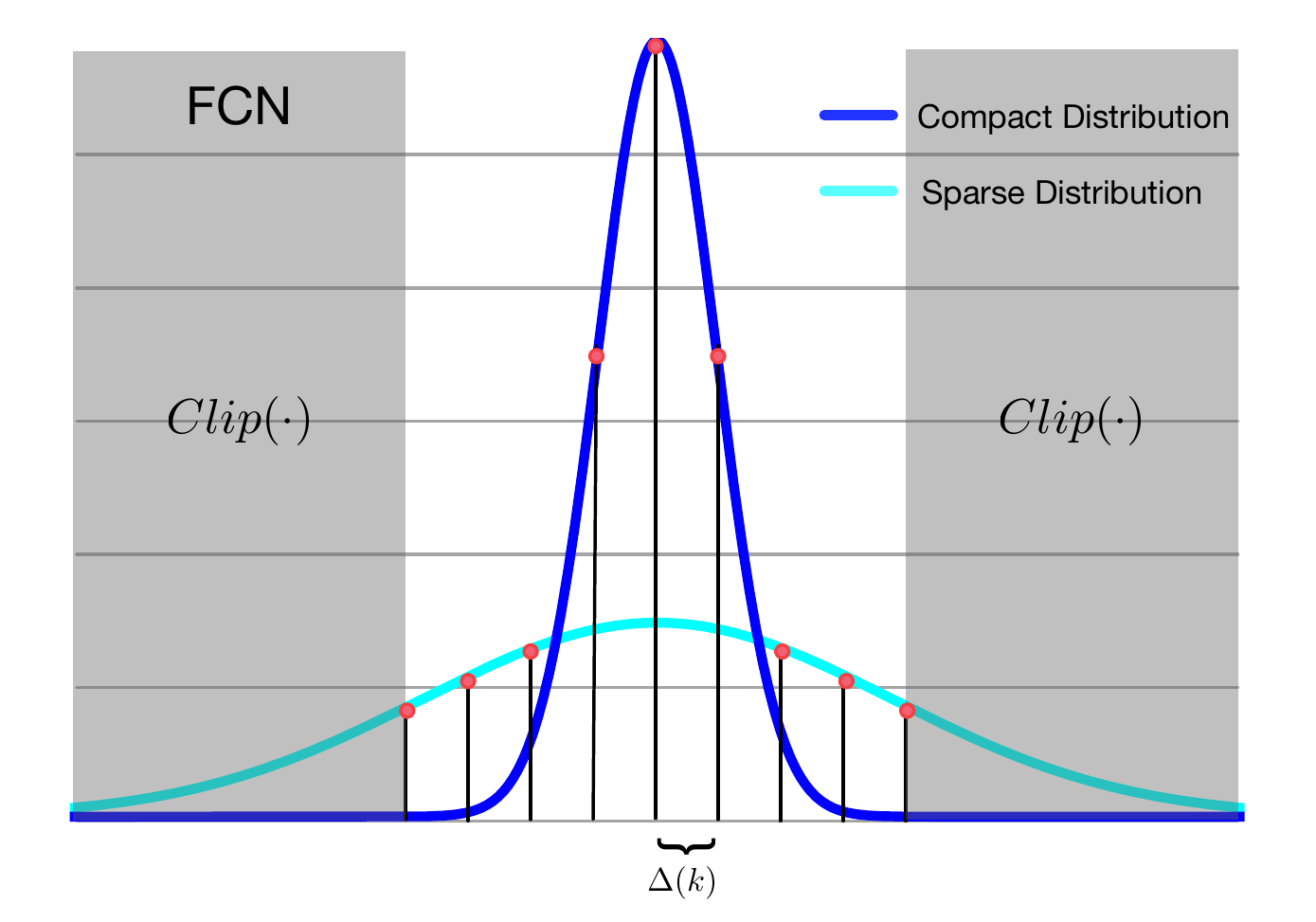}
 \caption{}
 \label{distri_FCN}
 \end{subfigure}\hfil
 \begin{subfigure}{0.5\linewidth}
 \includegraphics[width=\linewidth]{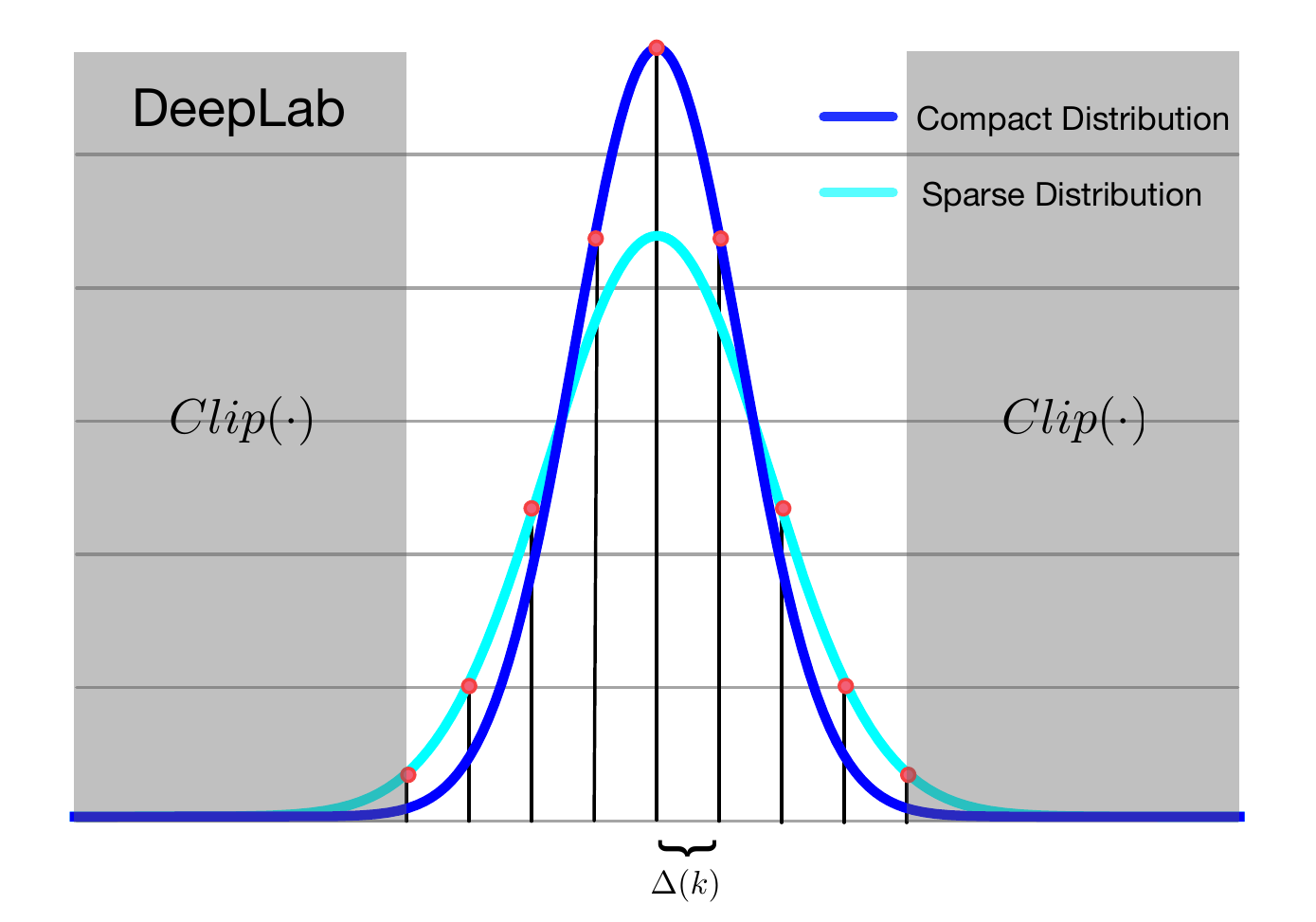}
 \caption{}
 \end{subfigure}\hfil
 \caption{Rough comparison of gradient distribution in FCN and DeepLab. Deep blue distribution describes the most compact distribution of a gradient layer, light blue describes the most scattered distribution of a gradient layer. (a) Gradient distribution of FCN, where gradients of different layers can have huge distribution differences. (b) Gradient distribution of DeepLab, where the distribution of different layers has a relatively smaller difference. }
 \label{Distri_comp}
\end{figure}

\section{Analysis on Back Propagation Quantization}\label{backpropA}
Backpropagation is the key to the convergence of neural networks, and different quantization methods can dramatically influence the result of semantic segmentation networks. This section aims to provide insights and specify different quantization methods based on the properties brought by semantic segmentation networks.

We have discussed previously the importance of dynamic scaling in objects that do not fit the uniform quantization range, but there still lies a problem on whether to keep the original order of magnitude of that layer, corresponding to Equation (\ref{GQ1}) and Equation (\ref{GQ2}). In previous works, DoReFa \cite{zhou2016dorefa} scales the value back to its original magnitude after quantization, and some other frameworks do not. WAGE \cite{wu2018training} omitted the scaling back based on the observation that it is the direction of error, rather than the order of magnitude leads the network towards convergence. This claim is partially valid depending on different contexts since their gradient is also quantized with dynamic shifting, therefore canceling out the scale influence in error quantization.

Keep in mind that our quantization framework simulates full precision training, and full precision networks have different magnitude of gradient in different layers. When not scaling back to the original order of magnitude, we suffer from information loss of the original distribution. Our experiment results in Section \ref{ExperimentSec} show that this information loss result in a noticeable degradation of accuracy in our DeepLab model.

However, this performance gain does not apply to all cases. For models that do not implement BN, the original order of magnitude should not be preserved. While the magnitude of gradient in these networks still varies, some extreme distribution in the decoder can have very large or small scale factors, resulting in a large dynamic range. Figure \ref{Distri_comp} roughly demonstrates the distribution of network with BN and network without BN in semantic segmentation. In a non-BN network, both failure modes are met mentioned in Section \ref{SQsection} due to the large dynamic range. This makes it hard to fit the quantization range of weight update quantization described in Equation (\ref{UQ2}). We discover that the extreme value in a non-BN network is introduced by the unique initialization method in semantic segmentation networks: First, the encoder uses pre-trained weights from DCNN on object classification tasks. Second, the full convolution layer and decoder layers are initialized into filters like bilinear upsampling. These two differences contribute to the large dynamic range of gradients in non-BN networks. Thus, dynamic scaling in these networks acts as a normalizing method in backpropagation. Although sacrificing the absolute magnitude of gradient in each layer, it helps to ensure valid weight updates in each training step.

\section{Experiments and Results}
To test the effectiveness of the quantization framework on semantic segmentation networks, we evaluate on two mainstream networks: FC8s-VGG16 \cite{long2015fully} and DeepLabv3-ResNet-50 \cite{DBLP:journals/corr/ChenPSA17}. The proposed models are trained and evaluated on PASCAL VOC 2012 and ADE20K datasets. PASCAL VOC 2012 includes 20 foreground object classes and one background class. The proposed models in this paper are trained using the augmented dataset provided by \cite{BharathICCV2011}, resulting in 10,582 training images, 1449 validation images, and 1456 test images.
ADE20K consists of 150 foreground object classes and no background class. The dataset contains 20210 training images and 2000 validation images. Since we aim at comparing quantized network against full precision network, all performance reported are averaged over ten run on the validation set to avoid randomness. 

The quality of the results is measured using mean Intersection over Union (mIoU). The computation formula of a single class IoU is governed by
\begin{equation}\label{iou}
IoU_i = \frac{TP_i}{TP_i + FP_i + FN_i}
\end{equation}
where for any class $i$, $TP_i$, $FP_i$, $FN_i$ denotes the pixels correctly predicted as class $i$, the pixels wrongly predicted as class $i$, and the pixels that are wrongly predicted as non-$i$ class, respectively. Then the mIoU is average on all appeared class in the evaluated image, we have
\begin{equation}\label{miou}
mIoU_i = \frac{1}{m}\sum_{j=1}^m IoU_j.
\end{equation}

\subsection{Implementation Details}
\textbf{FCN8s:} We use pre-trained weights for the VGG16 network backbone trained on Imagenet, and initialize the transpose convolution decoder to a bilinear upsampling filter with trainable weights. Every layer is fine-tuned with stochastic gradient descent. The network is trained with stochastic gradient descend optimizer with learning rate decaying to half for every 10 thousand steps. We use MSE loss function together with L2 weight decay of 5e-4. We crop the images into $512 \times 512$ with batch size of 8. Quantization bit-width are set to $k_W = k_A = k_{E_1} = k_G = 8$, and $k_U = 24$, and quantized gradients are scaled in each layer according to Equation (\ref{GQ2}).

\textbf{DeepLabv3:} The network backbone is initialized with ResNet-50 pre-trained weights trained on ImageNet, and the rest of the network are initialized using the MSRA method \cite{he2015delving}. Each layer is fine-tuned using softmax cross entropy with weight decay of 1e-5. We also use input images cropped at $512 \times 512$ and batch size of 8. Quantization bit-width are set to $k_W = k_A = k_{E_1} = k_G = 8, k_U = 24$ and no gradients scaling is performed during quantization, according to Equation (\ref{GQ1}). For BN quantization, $k_\mu = k_\sigma = k_{\hat{x}} = 16$ and $k_\gamma = k_\beta = 8$. For $k_{E_2}$, we report the results on both 8 bits and 16 bits.

\begin{table}
\centering
\caption{mIoU and pixel accuracy of different bit-width on FCN and DeepLab.}
\label{accuTable}
\normalsize
\begin{tabular}{c|ccccccc|cc} 
\hline
Model & $W$ & $A$& $G$ & $E_1$ & $E_2$ & $U$ & $BN$ &PASCAL & ADE20K\\  
\hline
\multirow{3}*{FCN}&32 & 32 & 32 & 32 & - & 32 &-& 65.17\% & 27.85\%\\  
&8 & 8 & 8 & 8 & - & 24 &-& \textbf{ 63.12\%} & \textbf{26.08\%} \\  
&4 & 4 & 8 & 8 & - & 24 &-& 34.68\% & 13.27\% \\  
\hline
\multirow{5}*{DeepLab}&32 & 32 & 32 & 32 & 32 & 32 &$L2BN$& 65.62\% & 26.14\%\\  
&8 & 8 & 8 & 8 & 16 & 24 &$L2BN$& 65.13\% & 25.92\% \\  
&8 & 8 & 8 & 8 & 16 & 24 &$L1BN$& \textbf{65.21\%} & \textbf{25.83\%} \\  
&8 & 8 & 8 & 8 & 8 & 24 &$L1BN$& 63.55\% & 24.16\% \\  
&4 & 4 & 8 & 8 & 8 & 24 &$L1BN$& 47.38\% & 16.69\% \\  
\hline
\end{tabular} 
\end{table}  

\begin{figure}[H]
 \centering
 \begin{subfigure}[b]{0.2\linewidth}
 \includegraphics[width=\linewidth]{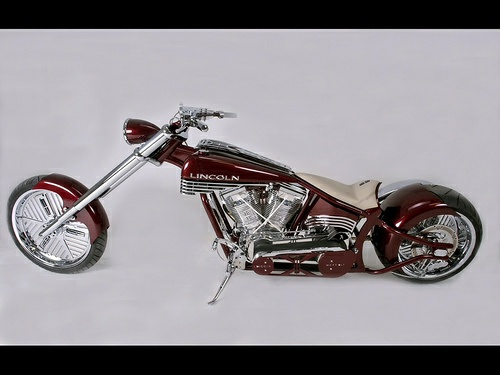}
 \end{subfigure}\hfil
 \begin{subfigure}[b]{0.2\linewidth}
 \includegraphics[width=\linewidth]{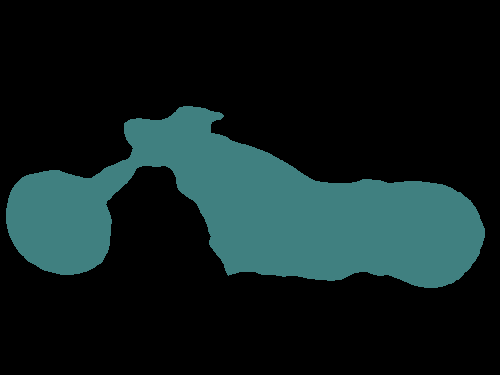}
 \end{subfigure}\hfil
 \begin{subfigure}[b]{0.2\linewidth}
 \includegraphics[width=\linewidth]{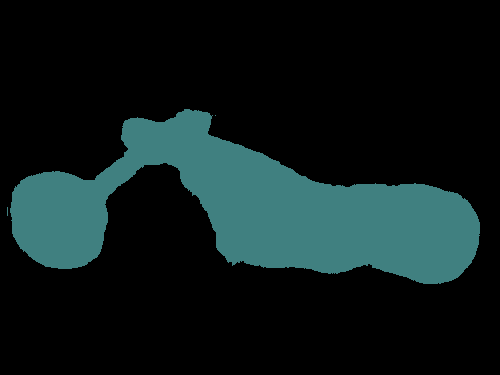}
 \end{subfigure}\hfil
  
 \medskip
 \begin{subfigure}[b]{0.2\linewidth}
 \includegraphics[width=\linewidth]{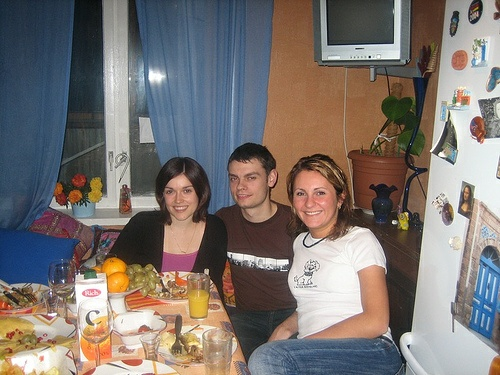}
 \end{subfigure}\hfil
 \begin{subfigure}[b]{0.2\linewidth}
 \includegraphics[width=\linewidth]{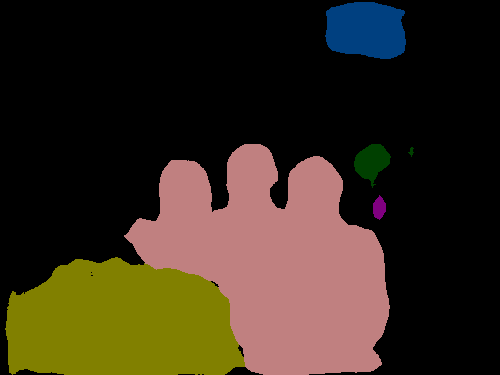}
 \end{subfigure}\hfil
 \begin{subfigure}[b]{0.2\linewidth}
 \includegraphics[width=\linewidth]{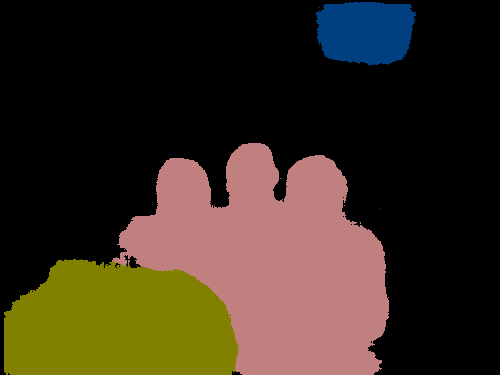}
 \end{subfigure}\hfil

 \medskip
 \begin{subfigure}[b]{0.2\linewidth}
 \includegraphics[width=\linewidth]{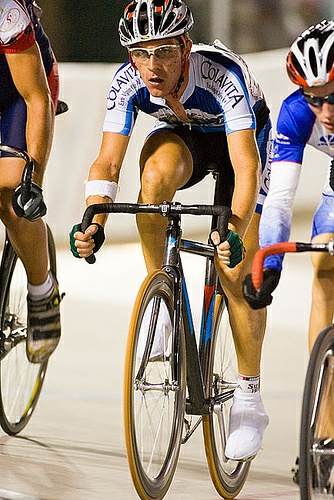}
 \caption{Original Image}
 \end{subfigure}\hfil
 \begin{subfigure}[b]{0.2\linewidth}
 \includegraphics[width=\linewidth]{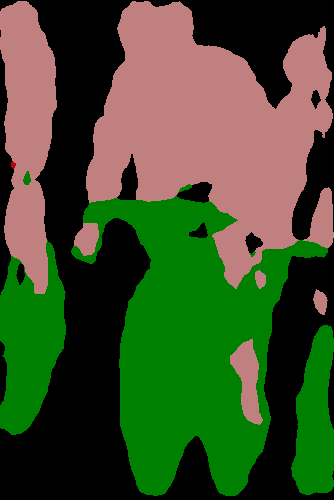}
 \caption{Full Precision}
 \end{subfigure}\hfil
 \begin{subfigure}[b]{0.2\linewidth}
 \includegraphics[width=\linewidth]{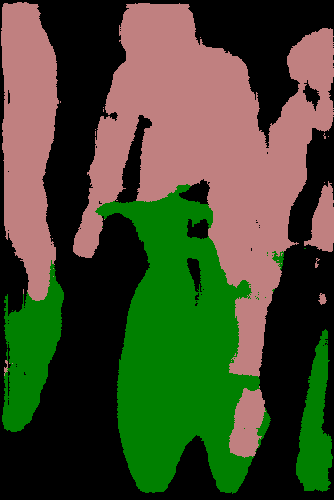}
 \caption{Quantized}
 \end{subfigure}\hfil
 \caption{Comparison of visual semantic segmentation results on full precision network and quantized network.}
 \label{quality_result}
\end{figure}

\subsection{Loss Curves and Accuracy}
In this section, we demonstrate the consistency of our framework on semantic segmentation networks. The quantitive results on FCN and DeepLab are summarized in Table \ref{accuTable}, where mIoU is recorded on both datasets. All hyperparameters except learning rate are kept the same for full-precision network and quantization network, and no techniques to improve accuracy such as data augmentation or test time augmentation has been used. The results we have achieved on quantization network is comparable to its full-precision counterpart. FCN reports 2.05\% and 1.77\% mIoU degradation on the default 8-bit quantization scheme compared with its full-precision counterpart on PASCAL and ADE20K, respectively. Quantized DeepLabv3 achieved precision on PASCAL with only 0.41\% degradation on the default scheme, and suffers 2.07\% degradation with $k_{E_2}$ also restricting to 8-bit quantization. Further, $L1BN$ proves to be equivalent to $L2BN$ in quantization network in terms of performance, but the linear property it preserves leaves the potential for implementation on integer-based hardware. Figure \ref{quality_result} demonstrates the comparison of qualitative results on full-precision network and quantized network.

In addition to the primary goal to achieve 8-bit quantization, we performed experiments with lower bit-width in forward and backward propagation to explore the potential in bit-width. We can see in Table {\ref{accuTable}} that the accuracy in both DeepLab and FCN suffers from noticeable accuracy loss when $k_W$ and $k_A$ are confined to 4-bit. This accuracy drop is primarily caused by the reason discussed in 3.1.2, where the sparse quantization levels generate a significant quantization error that affects the forward propagation of the network. Bit-width lower than 8-bit in backward propagation is not feasible, since the backward propagation directly impacts the weight update, resulting in a more adverse impact on performance compared to forward propagation. Hence, lower bit-width training requires different quantization approaches such as the BNN {\cite{DBLP:journals/corr/CourbariauxB16}} for better performance.

\begin{figure}
 \centering
 \begin{subfigure}{0.45\linewidth}
 \includegraphics[width=\linewidth]{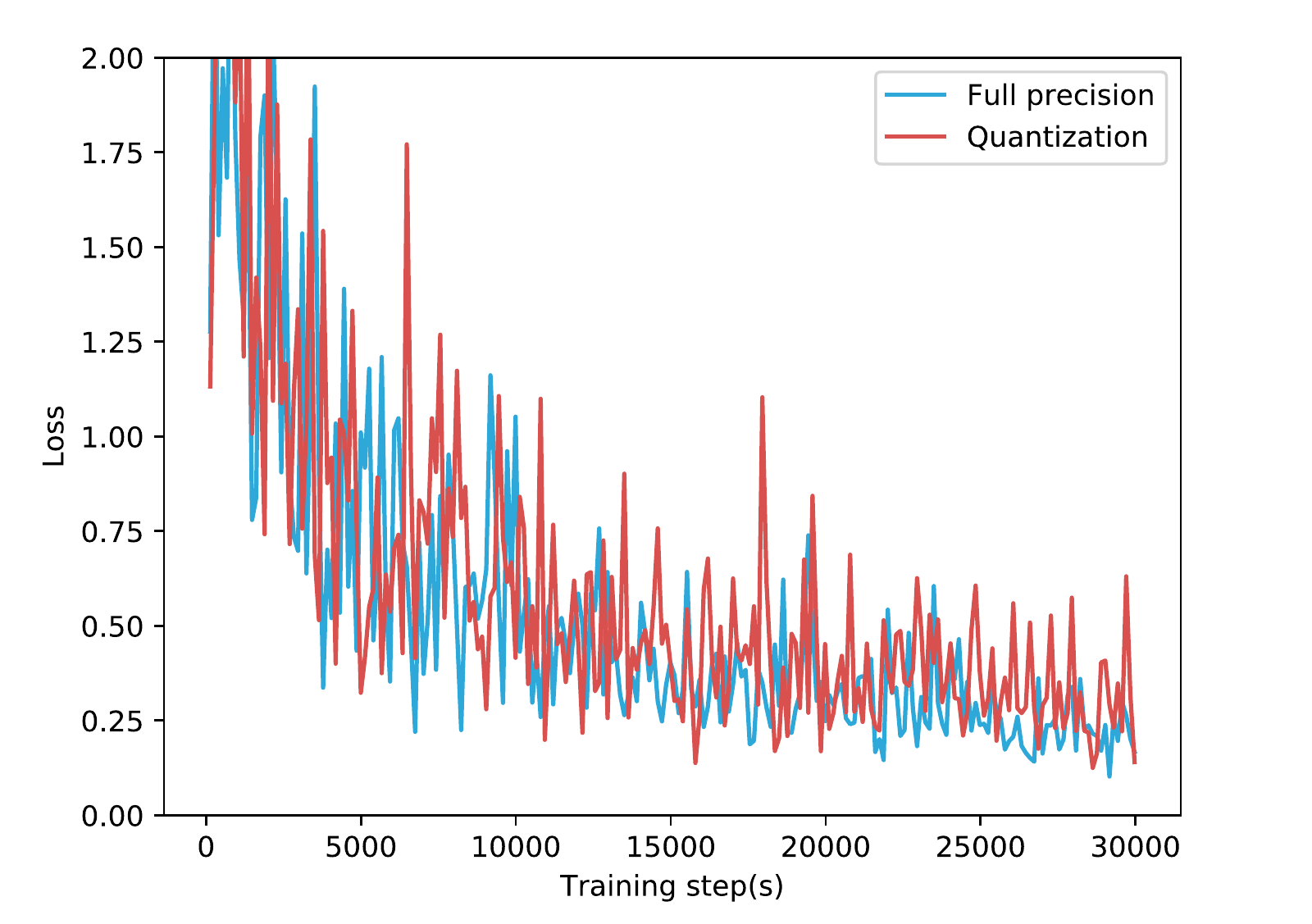}
 \caption{}
 \end{subfigure}
 \medskip
 \begin{subfigure}{0.45\linewidth}
 \includegraphics[width=\linewidth]{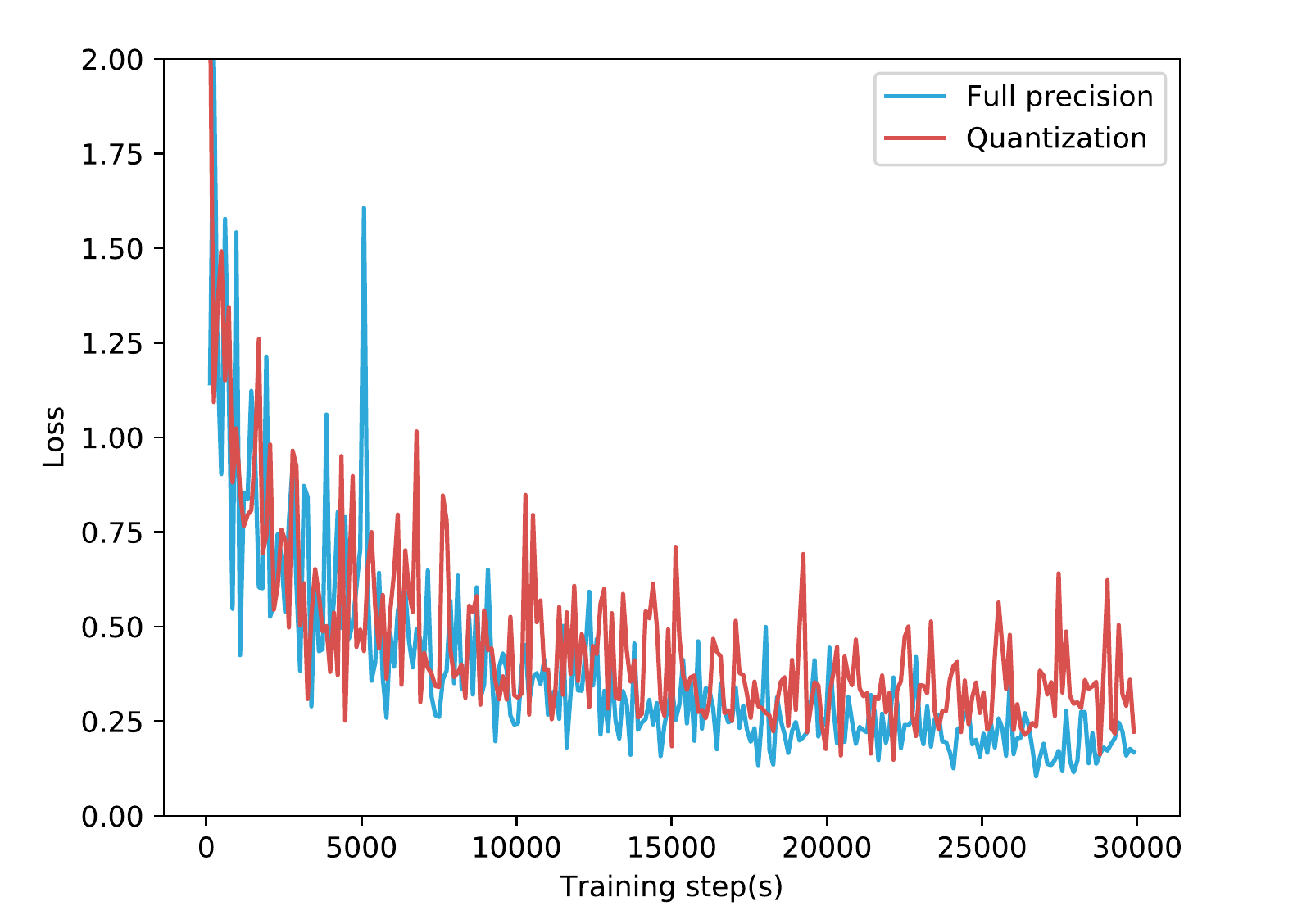}
 \caption{}
 \end{subfigure}
 \caption{Loss curve comparison between full precision network and standard quantized network: (a) FCN, (b) DeepLab.}
 \label{loss_curve}
\end{figure}

\indent Figure \ref{loss_curve}. shows the loss curve during training on PASCAL VOC 2012 data set. The quantized training curve follows the full-precision curve closely. This shows that our quantization framework effectively trained the network on discrete dataflow. FCN converges slower than DeepLab, and the curve tremors more seriously when quantized. This is due to the lack of normalization methods in FCN. The loss curve of quantized DeepLab converges similarly to the full precision training in early stages, but failed to close the gap at the final stages of training, and this gap is caused by the quantization step of weight update. With given quantization step, weight updates in the final training stage suffer from serious quantization noise, which can be resolved by increasing the bit-width of weight update.

\begin{table}[H]
\centering
\caption{Performance comparison of different quantization methods. Using scale means that the original magnitude is not preserved.}
\label{scaleAccu}
\normalsize
\begin{tabular}{c|cc|c} 
\hline
Model & Scale & $BN$ & mIoU \\  
\hline
\multirow{2}*{FCN}&\cmark&-& 63.12\% \\  
&\xmark&-& 33.32\% \\  
\hline
\multirow{2}*{DeepLab}&\cmark&L1BN& 64.13\% \\  
&\xmark&L1BN& 65.21\% \\  
\hline
\end{tabular} 
\end{table} 

\begin{figure}
 \centering
 \includegraphics[width=0.6\linewidth]{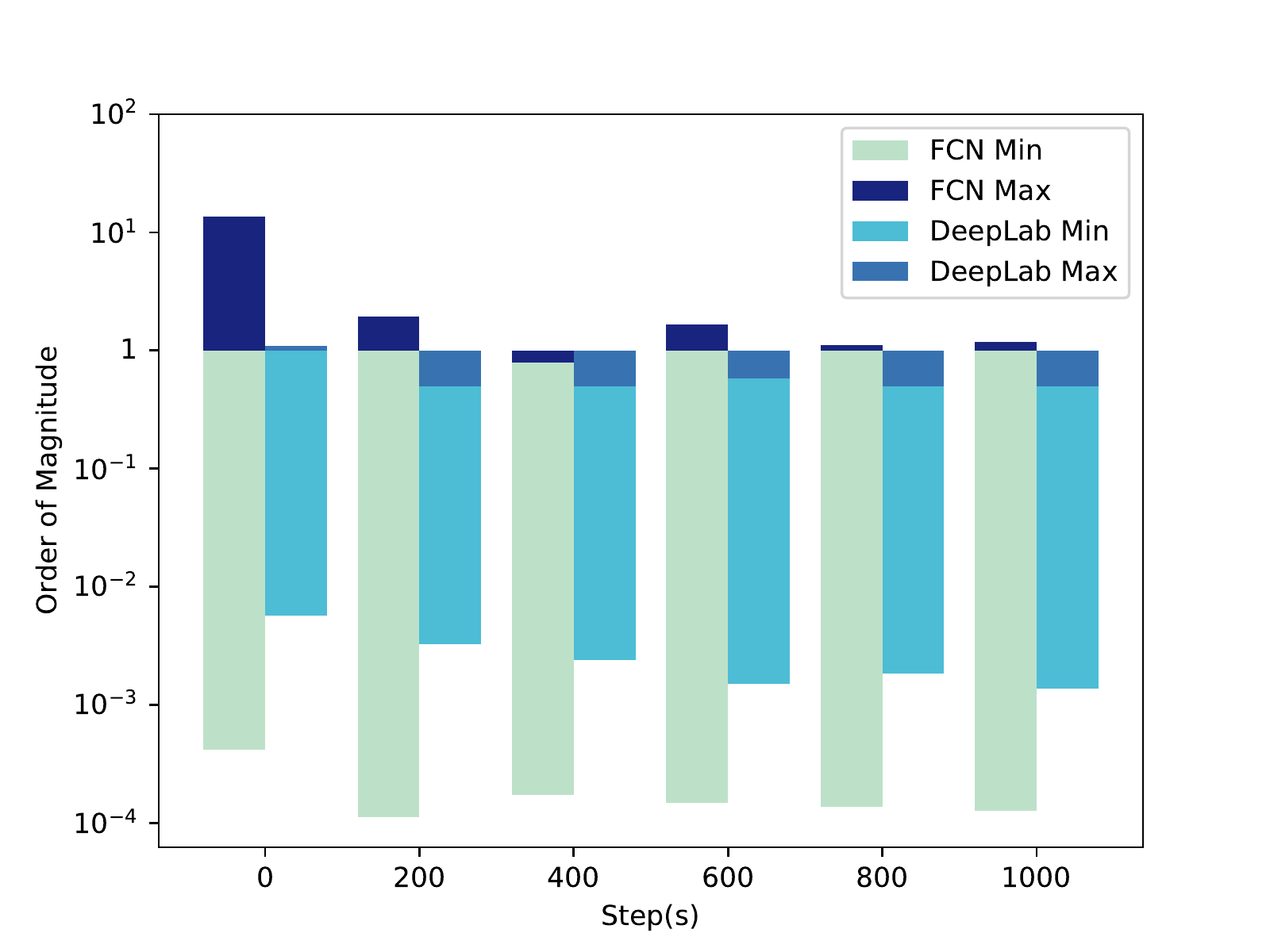}
 \caption{Maximum scale and minimum scale on FCN and DeepLab on different training stages. FCN has a larger difference compared with DeepLab in concern of gradient magnitudes.}
 \label{gradient_distri}
\end{figure}

\subsection{Performance on different scaling methods}\label{ExperimentSec}
In Section \ref{backpropA}, we theoretically analyzed the impact of preserving the original magnitude of gradients in backpropagation. Table \ref{scaleAccu} shows the accuracy of scaling the gradients or preserving the gradients. FCN suffers from huge mIoU degradation when preserving the original order of magnitude; Whereas DeepLab achieved 1.08\% improvement. Although the performance improvement is not huge, DeepLab manages to close off some gap of mIoU between full-precision and quantization network. To further analysis the different behavior on FCN and DeepLab, we examined the performance of both networks during the training process, and discovered that the first 1000 steps are crucial to training the networks on PASCAL VOC 2012 dataset, and later training steps try to fine-tune the network to its performance limit. Therefore we extracted the gradient distribution of the first 1000 steps of the training process, and found different behavior on DeepLab and FCN.

Figure \ref{gradient_distri} shows the maximum and minimum order of magnitude extracted from the gradient distributions in the network. We observe that the order of magnitude of each layer can have a huge difference between layers. The range of the order of magnitude is wide for the layers in FCN. The maximum order of magnitude comes from the decoder gradients and minimum from the pre-trained encoder, covering the range of $[10^{-4},10^{1}]$. On the other hand, DeepLab gradients order of magnitude is constrained around $[10^{-2},10^{0}]$. While the extreme distribution in FCN does not affect the full precision network, the quantized network is restricted from the quantization range and minimum quantization step, as presented in Figure \ref{distri_FCN}. This suggests that despite the slower convergence of the decoder, scaling the gradients to similar distribution in each layer ensures valid network updates in each step. Since DeepLab has a more compact distribution of gradients, by not scaling the gradients prevent the network from losing distribution information during training, resulting in a slight increase in performance. 

\subsection{Bit-width impact on weight update and weights}
Bit-width of weight update $k_W$ is strongly correlated to the performance of our network, and usually takes more bits compared to other quantization objects for the model to converge properly. As shown in Equation (\ref{UQ2}), weight update determines the bit-width of master weights. Figure \ref{ubit_curve} demonstrates the relationship between $k_U$ and network performance, keeping other quantization parameters the same as the default bit-width. We found that as $k_U$ approaches $k_W$, the network performance drops dramatically. Both DeepLab and FCN still generate acceptable results when $k_U$ is above 16 bits, thus weight update bit-width can be further compressed to save storage space. As $k_U$ approaches the lower bound of $k_W$, the fluctuation of weights increases, and performance quickly degenerates to 1\%. This is because when $k_U$ is bigger, master weights create a buffer for quantized weights, so only the accumulated valid weight updates take effect finally in quantized weights.

\begin{figure}
 \centering
 \includegraphics[width=0.6\linewidth]{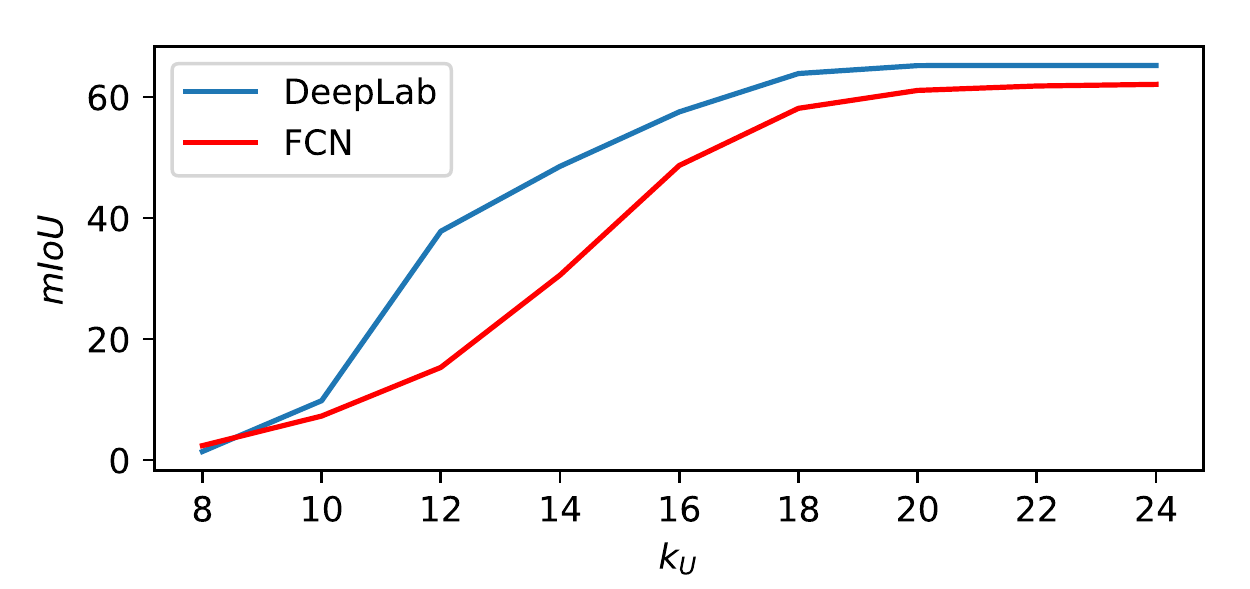}
 \caption{mIoU performance curve of DeepLab and FCN on different $k_U$ on PASCAL VOC 2012. Both models suffer from noticeable performance loss when $k_U$ is lower than 16 bits. }
 \label{ubit_curve}
\end{figure}

\subsection{Sensitivity of Decoder}
The decoder is a critical part of semantic segmentation network that differs with traditional DCNN. While extracting features, feature maps are scaled to smaller sizes during pooling operation. Hence, the decoder recovers the resolution using layers like deconvolution layer and produces pixel-wise outputs. Here we use the FCN model to study the impact of decoder on our quantized framework (note that in experiments we divide the full convolution layer to the decoder structure for simplicity, since it does not belong to traditional DCNN).

\begin{table}
\centering
\caption{Performance comparison of quantizing different parts of dataflow. Decoder is more sensitive to quantization, responsible for a large portion of performance degradation.}
\label{endeq}
\normalsize
\begin{tabular}{cc|c} 
Encoder &Decoder & mIoU \\  
\hline
FP32&FP32& 65.17\% \\  
\textbf{INT8}&FP32&64.74\% \\  
FP32&\textbf{INT8}& 62.95\% \\  
\textbf{INT8}&\textbf{INT8}&63.12\% \\  
\end{tabular} 
\end{table} 

\begin{figure}[H]
 \centering
 \begin{subfigure}[b]{0.3\linewidth}
 \includegraphics[width=\linewidth]{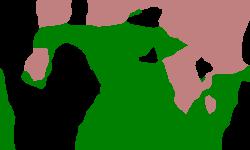}
 \caption{Full Precision}
 \end{subfigure}\hfil
 \begin{subfigure}[b]{0.3\linewidth}
 \includegraphics[width=\linewidth]{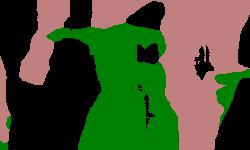}
 \caption{Encoder Quantized}
 \end{subfigure}\hfil
  
 \medskip
 \begin{subfigure}[b]{0.3\linewidth}
 \includegraphics[width=\linewidth]{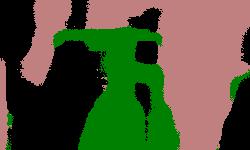}
 \caption{Decoder Quantized}
 \end{subfigure}\hfil
 \begin{subfigure}[b]{0.3\linewidth}
 \includegraphics[width=\linewidth]{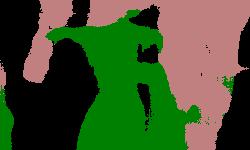}
 \caption{Quantized}
 \end{subfigure}\hfil
 \caption{Visual results of segmentation prediction on quantizing different structures of the network.}
 \label{data_q_compare}
\end{figure}

To analyze the impact of quantizing decoder, we performed an additional two set of experiments that solely quantizes either the encoder or decoder, and other parameters strictly follows the default quantization scheme. Further, we compare them with the results produced with full precision training and quantized training results acquired earlier, which is shown in Table \ref{endeq}.

The results suggest that the decoder seems to contribute more to the performance degradation compared to the decoder in quantization. To better understand this degradation, we extracted visual results from these four experiments and compared them. Figure \ref{data_q_compare} shows one set of the qualitative results, which represents the general failure mode of decoder well. While full precision and encoder quantized network produces results with a smooth boundary of segmentation mask, the boundary of decoder quantized network and entirely quantized network(default scheme) produces results with a rough edge when zooming the results. Moreover, we observe that a portion of the labeled pixels even separates from its boundary. As the overall visual segmentation results between these networks are similar, we suspect that the $2.05\%$ performance largely comes from the rough edge.

\begin{figure}
 \centering
 \includegraphics[width=0.7\linewidth]{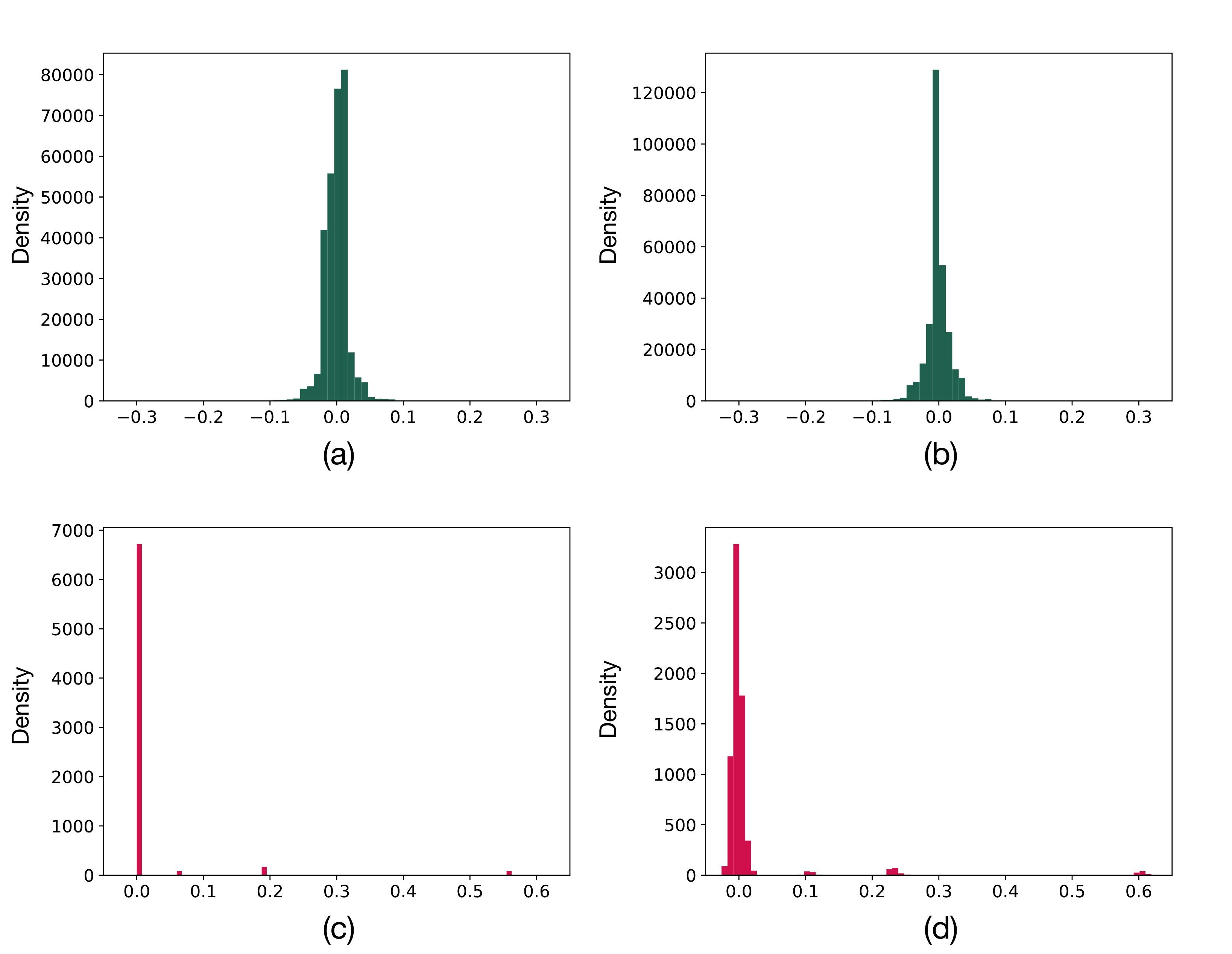}
 \caption{Comparison of weight distribution on early and final training stages of the decoder in FCN. (a) Encoder early stage, (b) Encoder final stage, (c) Decoder early stage, (d) Decoder final stage.}
 \label{decoder_weights}
\end{figure}

Prior experience tells us that the quality of the decoder is closely related to the boundaries between different segmentation objects. Lead by this thought, we extracted the weights in the deconvolution layer in the early training stage and final training stage, shown in Figure \ref{decoder_weights}. Compared with the compact distribution of encoder variables, most values in the decoder are small values around zero, with sparse bigger values scattered in the range. This distribution suggests larger quantization noise compared to the distribution of traditional convolution layers. Due to this reason, quantization on the decoder cannot be quantized to lower than 8 bits, or the quantization noise will seriously affect backpropagation, leading the network to diverge. Although dynamic scaling quantization may not help much in quantizing the decoder, future work can try adaptive quantization to close the slight accuracy gap between full precision network and quantized network.

\section{Conclusion and Future Work}
In this paper, we thoroughly analyzed network quantization in semantic segmentation and presented a quantization framework that achieves comparable performance on public datasets. It is promising that performance can be further improved on the quantization of decoder, by designing adaptive quantization methods to reduce the quantization error during decoder backpropagation. As it is not specifically designed for any specific network architecture, our quantization framework can be adapted to other semantic segmentation networks and provide acceleration in computation speed and save storage space on specific hardware. The fully quantized dataflow and removal of the square and root operations in batch normalization promise the future for efficient online training and integer-based deep learning accelerators. 

Despite the efforts made in this paper, there is still work to do in BN quantization. Existing integer-based deep learning accelerators fall short of computing the mean and variance in BN. Recent works like show that BN can be removed from full-precision network by carefully initializing weights, but the result of directly implementing it on a quantized network is not promising. This may be because quantized networks is very sensitive to zero-initialized values, making it very hard for the weight to properly update at early training stages. Future work can focus on various ways of removing BN from the network, and leave only weight ($W$), activation ($A$), error ($E$), gradient ($G$) and update ($U$) in the quantization framework. With BN removed, quantization for semantic segmentation can be more efficient and easier to implement on integer-based hardware.

\section*{Acknowledgement}
The work was partially supported by Tsinghua University Initiative Scientific Research Program, and Tsinghua-Foshan Innovation Special Fund (TFISF), and National Science Foundation of China (61876215).

\section*{References}

\bibliography{elsarticle-template}

\end{document}